\documentclass{article}

\usepackage[final]{nips_2018}
\usepackage[utf8]{inputenc} % allow utf-8 input
\usepackage[T1]{fontenc}    % use 8-bit T1 fonts
\usepackage{hyperref}       % hyperlinks
\usepackage{url}            % simple URL typesetting
\usepackage{booktabs}       % professional-quality tables
\usepackage{amsfonts}       % blackboard math symbols
\usepackage{nicefrac}       % compact symbols for 1/2, etc.
\usepackage{microtype}      % microtypography
\usepackage{xcolor}         % for comments

\usepackage{natbib}
\title{Dialog System Technology Challenge 7}

\author{
  Koichiro Yoshino, Chiori Hori, Julien Perez, Luis Fernando D'Haro, \\
  {\bf Lazaros Polymenakos, Chulaka Gunasekara, Walter S. Lasecki,} \\
  {\bf Jonathan K. Kummerfeld, Michel Galley, Chris Brockett, Jianfeng Gao, }\\
  {\bf Bill Dolan, Xiang Gao, Huda Alamari,
  Tim K. Marks, Devi Parikh and Dhruv Batra\thanks{Every author has equal contribution. http://workshop.colips.org/dstc7}}
}

\begin{document}
% \nipsfinalcopy is no longer used

\maketitle

\begin{abstract}

This paper introduces the Seventh Dialog System Technology Challenges (DSTC), which use shared datasets to explore the problem of building dialog systems.
Recently, end-to-end dialog modeling approaches have been applied to various dialog tasks.
The seventh DSTC (DSTC7) focuses on developing technologies related to end-to-end dialog systems for (1) sentence selection, (2) sentence generation and (3) audio visual scene aware dialog.
This paper summarizes the overall setup and results of DSTC7, including detailed descriptions of the different tracks and provided datasets.
We also describe overall trends in the submitted systems and the key results.
Each track introduced new datasets and participants achieved impressive results using state-of-the-art end-to-end technologies.

\end{abstract}

\section{Introduction}

The ongoing DSTC series started as an initiative to provide a common testbed for the task of Dialog State Tracking; the first edition was organized in 2013 (\cite{williams2013dialog}) and used human-computer dialogs in the bus timetable domain.
Dialog State Tracking Challenges 2 (\cite{henderson2014second}) and 3 (\cite{henderson2014third}) followed in 2014, using more complicated and dynamic dialog states for restaurant information in several situations: dialog state tracking for unseen states and different domain data from the training data.
Dialog State Tracking Challenge 4 (\cite{kim2017fourth}) and Dialog State Tracking Challenge 5 (\cite{kim2016fifth}) moved to tracking human-human dialogs in mono- and cross-language settings.
For the most recent event, DSTC 6 in 2017, the acronym was changed to mean Dialog System Technology Challenge (\cite{hori2018overview}) and focused on end-to-end systems with the aim of minimizing effort on human annotation while exploring more complex tasks. 

As we can see, since 2013 the challenge has evolved in several ways.
First, from modeling human-computer interactions, to investigating human-human interactions, and finally moving toward complex end-to-end systems.
DSTC has also offered pilot tasks on Spoken Language Understanding, Speech Act Prediction, Natural Language Generation and End-to-End System Evaluation, which expanded interest in the challenge in the research communities of dialog systems and AI.
Therefore, given the remarkable success of the first five editions, the complexity of the dialog phenomenon and the interest of the research community in the broader variety of dialog related problems, the DSTC rebranded itself as "Dialog System Technology Challenges" for its sixth edition. 

For the seventh event, there were five task proposals.
These were discussed at the sixth event, with a particular focus on how applied proposals were, and how they fit within the larger space of problems of interest to the research community.
Three critical issues were raised in the discussion.
First, the retrieval-based approach for response generation is still essential for practical use, even if the generative approach often used by neural conversation models has had enormous success (Sentence Selection Track).
Second, working on improving generative approaches is also important, but results generated by systems should have more variety according to their contexts, including dialog histories, locations, and other dialog situations (Sentence Generation Track).
The final issue is fusion with other areas; visual dialog is one direction in which information in images is ued in the dialog (Audio Visual Scene-Aware Dialog Track).
Following the discussion, three tasks were proposed for the seventh dialog system technology challenge, as described below.

In Sentence Selection (described in more detail in section \ref{sentence_selection}), the challenge consists of several sub-tasks, in which system are given a partial conversation, and they must select the correct next utterances from a set of candidates or indicate that none of the proposed utterances is correct.
This is intended to push the utterance classification task towards real-world problems. 

In Sentence Generation (described in detail in section \ref{sentence_generation}), the goal is to generate conversational responses that go beyond chitchat, by injecting informational responses that are grounded in external knowledge.
Since there is no specific or predefined goal, this task does not constitute what is commonly called task-oriented dialog, but target human-human dialogs where the underlying goal is often ill-defined or not known in advance. 

Finally, in the Audio Visual Scene-aware track (described in detail in section \ref{avsd}), the goal is to generate system responses in a dialog about an input video.
Dialog systems need to understand scenes to have conversations with users about the objects and events around them.
In this track multiple research technologies are integrated, including: end-to-end dialog technologies, which generate system responses using models trained from dialog data; visual question answering (VQA) technologies, which answer to questions about images using learned image features; and video description technologies, in which videos are described/narrated using multimodal information.  

\section{Sentence Selection Track}
\label{sentence_selection}
This task\footnote{https://ibm.github.io/dstc7-noesis/public/index.html} pushed the state-of-the-art in goal-oriented dialog systems in four directions deemed necessary for practical automated agents, using two new datasets.
We sidestepped the challenge of evaluating generated utterances by formulating the problem as response selection, as proposed by \citet{lowe-EtAl:2015:W15-46}.
At test time, participants were provided with partial conversations, each paired with a set of utterances that could be the next utterance in the conversation.
Systems needed to rank these options, with the goal of placing the true utterance first.
Unlike prior work, we considered several advanced variations of the task:

\begin{description}
  \item[Subtask 1] 100 candidates, including 1 correct option.
  \item[Subtask 2] 120,000 candidates, including 1 correct option (Ubuntu data only).
  \item[Subtask 3] 100 candidates, including 1-5 correct options that are paraphrases (Advising data only).
  \item[Subtask 4] 100 candidates, including 0-1 correct options.
  \item[Subtask 5] The same as subtask 1, but with access to external information.
\end{description}

These subtasks push the capabilities of systems and enable interesting comparisons of strengths and weaknesses of different approaches.
Participants were able to use the provided knowledge sources as is, or automatically transform them to other representations (e.g. knowledge graphs, continuous embeddings, etc.) that would improve their dialog systems.

Comparing to the DSTC6 Sentence Selection track, this year's track differed in several ways.
Most importantly, we use human-human dialogs, rather than a synthetically created dataset.
Each of our subtasks also adds a novel dimension compared to the DSTC6 task, which provided candidate sets of size 10 with a single correct option, and no external resource.

\subsection{Data}

\begin{figure}
\centering
\begin{tabular}{ll}
Questioner: & how do I turn on optical output under gutsy?. (soundcard) \\
Helper: & probably check the settings in the mixer \\
Questioner: & I've tried that, speakers still say no incoming signal. \\
Helper: & there should be some check box for analog/digital output, \\
& but unfortunately I wouldn't know much more \\
\\
Student: & Hello! \\
Advisor: & Hello. \\
Student: & I'm looking for good courses to take. \\
Advisor: & Are you looking for courses in a specific area of CS? \\
Student: & Not in particular. \\
Advisor: & Are you looking to take a very difficult class? \\
\end{tabular}

    \caption{\label{fig:task1-example}
    Examples of partial dialogs in task one (Ubuntu top, Advising bottom).
    }
\end{figure}

Our datasets are derived from collections of two-party conversations.
The conversations are randomly split part way through to create a partial conversation and the true follow-up response.
Incorrect candidate utterances are selected by randomly sampling utterances from the dataset.
For the data with paraphrases, the incorrect candidates are sampled with paraphrases as well.
For the data where sometimes the pool does not contain the correct utterance, twenty percent of cases are selected at random to have no correct utterance.

This task considers datasets in two domains.
First, a collection of two-party conversations from the Ubuntu support channel, in which one user asks a question and another helps them resolve their problem.
These are extracted using the model described by \citet{Kummerfeld-EtAl:2018:arXiv}, instead of the heuristic approach used in \citet{lowe-EtAl:2015:W15-46}.
This approach produced 135,000 conversations, which we sample 100,000 of for training and 1,000 for testing.
For this setting, manual pages are provided as a form of knowledge grounding.

Second, a new collection of conversations in a student advising domain, where the goal is to help a student select courses for the coming semester.
These were collected at the University of Michigan with students playing both roles with simulated personas, including information about preferences for workloads, class sizes, topic areas, time of day, etc.
Both participants had access to the list of courses the student had taken previously, and the adviser had access to a list of suggested courses that the student had completed the prerequisites for.
In the shared task, we provide all of this information - student preferences, and course information - to participants.
815 conversations were collected, with on average 18 messages per conversation and 9 tokens per message.
This data was expanded by collecting 82,094 paraphrases of messages.
Of this data, 700 conversations were used in the shared task, with 500 for training, 100 for development, and 100 for testing.
The remaining 115 conversations were used as a source of negative candidates in the sets systems choose from.
For the test data, 500 conversations were constructed by cutting the conversations off at 5 points and using paraphrases to make 5 distinct conversations.
The training data was provided in two forms.
First, the 500 training conversations with a list of paraphrases for each utterance, which participants could use in any way.
Second, 100,000 partial conversations generated by randomly selecting paraphrases.

Finally, as part of the challenge, we provided a baseline system that implemented the Dual-Encoder model from \citet{lowe-EtAl:2015:W15-46}.
This lowered the barrier to entry, encouraging broader participation in the task.

\subsection{Results}

We considered a range of metrics when comparing models.
Following \citet{lowe-EtAl:2015:W15-46}, we use Recall@N, where we count how often the correct answer is within the top N specified by a system.
In prior work the set of candidates was 10 and N was set at 1, 2, and 5.
Since our sets are larger, we consider 1, 10, and 50.
We also consider a widely used metric from the ranking literature: Mean Reciprocal Rank (MRR).
Finally, for subtask 3 we use Mean Average Precision (MAP) since there are multiple correct utterances in the set.
To determine a single winner for each subtask, we used the mean of Recall@10 and MRR.

Twenty teams participated in at least one of the subtasks, seventeen participated in two or more, and three participated in every subtask.
For both datasets the subtask with the most entries was the first, which is closest to prior tasks.
One team had a clear lead, scoring the highest across all but one of the subtasks (task 2 on Ubuntu, when the number of candidates is increased).
The Advising data was consistently harder than the Ubuntu data, probably because of the limited training data.
However, the size of the Ubuntu dataset also posed a challenge in training, as substantial computation was required for even a single training epoch.

The best system had a Recall@1 of 0.645 on the first subtask for Ubuntu, and was based on the Enhanced Sequential Inference Model (ESIM) architecture proposed by \citet{chen2016enhanced}.
Their score on the second subtask was 0.067, which is a factor of ten lower, but with more than thousand times as many options to choose from.
The introduction of cases with no correct answer (subtask four) led to slightly lower results (0.511), while the availability of external data (subtask 5) helped slightly (0.653).
We see a similar trend on the Advising data, except that external data was less useful.

\subsection{Summary}
This track introduced two new dialog datasets to the research community and a range of variations on the sentence selection task.
The best submitted system managed to achieve Recall@1 score of 0.645 on Ubuntu, an impressive result given the large number of candidates and the complexity of the dialog.
One outstanding challenge is how to effectively use external information -- none of the teams managed to substantially improve performance from subtask 1 to subtask 5.

\section{Sentence Generation Track}
\label{sentence_generation}
% Prior work on fully data-driven:
Recent work \citep[etc.]{ritter2011data,sordoni2015,shang2015neural,vinyals2015neural,serban2015hierarchical} has shown that conversational models can be trained in a completely end-to-end and data-driven fashion, without any hand-coding.
However, prior work has mostly focused to chitchat, as that is a common feature of messages in the social media data (e.g., Twitter \citep{ritter2011data}) used to train these systems. 
To effectively move beyond chitchat and produce system responses that are both substantive and ``useful'', fully data-driven models need grounding in the real world and access to external knowledge (textual or structured).
To do so, the Generation Task of this year is inspired by the {\it knowledge-grounded} conversational framework of \cite{grounded2018}, which combines conversational input and textual data from the user's environment (here, a web page that is discussed).
Such a framework maintains the benefit of fully data-driven conversation while attempting to get closer to task-oriented scenarios, with the goal of informing and helping the users and not just entertaining them. 

\subsection{Task definition}

%We adopt an end-to-end conversational modeling task that is inspired by the framework of \cite{grounded2018}.
%Its goal was to generate conversational responses that go beyond chitchat by exploiting information drawn from external textual sources (in their case, Foursquare short reviews, otherwise known as ``tips'').
%Note that in this proposed track, there is no specific or predefined goal (e.g., booking a flight, or reserving a table at a restaurant), so this task does not constitute what is commonly called either goal-oriented or task-oriented dialog. 
%We do not see this as a limitation---even in human-human dialogs---the underlying goal is often ill-defined or not known in advance, even in work and other productive environments (e.g., brainstorming meetings).
%Overall, the task still 
The task
follows the data-driven framework established in 2011 by \cite{ritter2011data}, which avoids hand-coding any linguistic, domain, or task-specific information.
In the knowledge-grounded setting of \cite{grounded2018}, that framework is extended as each system input consists of two parts:\\
%
%\begin{itemize}[noitemsep]
%\item 
%{\bf Conversational input}: 
{\bf Conversational input:}
Similar to DSTC6 Track 2 \citep{DSTC6T2}, all preceding turns of the conversation are available to the system.
For practical purposes, we truncate the context to the $K$ most recent turns.\\
%\item  
%{\bf Contextually-relevant ``facts''}: 
{\bf Contextually-relevant ``facts'':} 
The system is given snippets of text that are relevant to the context of the conversation. 
These snippets of text are not drawn from any conversational data, and are instead extracted from external knowledge sources such as Wikipedia or Foursquare.
%\end{itemize}

From this input, the task it to produce a response that is both conversationally appropriate and informative.
The evaluation setup is presented in Section~\ref{sec:task2:eval}. 

\subsection{Data}

We extracted conversation threads from Reddit data, which is particularly well suited for grounded conversation modeling.
Indeed, Reddit conversations are organized around submissions, where each conversation is typically initiated with a URL to a web page (grounding) that defines the subject of the conversation.
For this task, we restrict ourselves to submissions that contain exactly one URL and a title.
To reduce spamming and offensive language and improve the overall quality of the data, we manually whitelisted the domains of these URLs and the Reddit topics (i.e., ``subreddits'') in which they appear.
This filtering yielded about 3 million conversational responses and 20 million facts divided into train, validation and tests.\footnote{We could have easily increased the number of web domains to create a bigger dataset, but we aimed to make the task relatively accessible for participants with limited computing resources.}
For the test set, we selected conversational turns for which 6 or more responses were available, in order to create a multi-reference test set.
Given other filtering criteria such as turn length, this yielded a 5-reference test set of size 2208 (For each instance, we set aside one of the 6 human responses to assess human performance on this task).
%Reddit provides an intuitive direct link to external data in the submissions that can be utilized for this task. 
More information about the data for this task can be found on the data extraction web site, which makes available all of the data extraction and evaluation code.\footnote{\url{https://github.com/DSTC-MSR-NLP/DSTC7-End-to-End-Conversation-Modeling}}

\subsection{Evaluation}
\label{sec:task2:eval}

We evaluate response quality using both automatic and human evaluation. 
Since we are not considering task-oriented dialog, there is no pre-specified task and therefore no extrinsic way of measuring task success. 
Instead, we performed a per-response human evaluation judging each system response using crowdsourcing:\\
%\begin{itemize}[noitemsep]
%\item 
%{\bf Appropriateness}: 
{\bf Relevance:}
This evaluation criterion asks whether the system response is conversationally appropriate and relevant given the $K$ immediately preceding turns (we set $K=2$ to reduce the judges' cognitive load).
%\footnote{We need to limit the number of preceding turns as to not overload MTurk workers with too much information.} 
Note that this judgment has nothing to do with grounding in external sources, and is similar to human judgments for prior data-driven conversation models (e.g., \citep{sordoni2015}).\\
%Nevertheless, this judgment is important, as we do not want systems to produce much less appropriate responses for the sake of being more informational.
%\item {\bf Informativeness \& Utility}: 
{\bf Interest:}
This evaluation criterion measures the degree to which the produced response is interesting and informative in the context of a document provided by the URL.
Since it would be impractical to show entire web pages to the crowdworkers, we restricted ourselves at training and test time to URLs with named anchors (i.e., prefixed with `\#' in the URL), and the crowdworkers only had to read a snippet of the document immediately following that anchor.
Note that models could use full web pages as input, and the decision to only show a snippet for each response was again to reduce cognitive load.
%\end{itemize}

We scored both evaluation criteria on a 5-point Likert scale, and finally combined the two judgments by weighting them equally. 
In order to provide participants with preliminary results to include in their system descriptions, we also performed automatic evaluation using standard machine translation metrics, including BLEU \citep{Papineni2002BLEU}, METEOR \citep{Lavie2007}, and NIST \citep{Doddington:2002}.
NIST is a variant of BLEU that weights $n$-gram matches by their information gain, i.e., it indirectly penalizes uninformative $n$-grams such as ``I don't'' and ``don't know''.
The final ranking of the systems was based only on human evaluation scores.

%\footnote{Those metrics would work as a proxy to the appropriateness judgment, not informativeness. We would use single-reference BLEU and METEOR, as Twitter provides one reference for each response.}

\subsection{Results}
\label{sec:task2:results}

The Generation Task received 26 system submissions from 7 teams.
In addition to these systems, we also evaluated a ``human'' system (one of the six human references set aside for evaluation) and three baselines: a seq2seq baseline, a random baseline (which randomly selected responses from the training data), and a constant baseline (which always responds ``I don't know what you mean.'').
The reason for including a constant baseline is that such a deflective response generation system can be surprisingly competitive, at least when evaluated on automatic metrics (BLEU).

The findings are as follows for each of the metrics:\\
{\bf BLEU-4:} When evaluated on 5 references, the constant baseline, which always responds deflectively, does surprisingly well (BLEU=2.87\%) and outperforms all the submitted systems (BLEU4 ranging from 1.01\% to 1.83\%), and is only outperformed by humans.
In further analysis, we found that reducing the number of references to one solved the problem, as almost all the systems were able to outperform the baseline according to single-reference BLEU.
We suspect this deficiency of {\it multi-reference} BLEU, previously noted in \cite{vedantam:15}, to be due to its parameterization as a precision metric.
For example, if one of the gold responses happens to be ``I don't know what you mean'', the constant baseline gets a maximum score for that instance, even if the other references are semantically completely unrelated.
Thus, this biases the metric towards bland responses, as often at least one of the 5 references is somewhat deflective (e.g., contains ``I don't know'').
Based on these observations, we chose to use single-reference BLEU instead of multi-reference BLEU for this DSTC task, as the former gave much more meaningful results. \\
{\bf NIST-4:} The NIST score weights ngram matches by their information gain, and effectively penalizes common $n$-grams such as ``I don't know'', which alleviates the problem with multi-reference BLEU mentioned above.
None of the baselines is competitive with the top systems according to NIST-4, even when using 5 references.
This suggests that NIST might be a more suitable metric than BLEU when dealing with multi-reference test sets, and it penalizes bland responses. \\
{\bf METEOR:} This metric suffers from the same problem as BLEU-4, as the constant baseline performs very well on that metric and outperforms all submitted primary systems but one.
We suspect this is due to the fact that METEOR (as BLEU) does not consider information gain in its scoring. \\
{\bf Human Evaluation:} Owing to the cost of crowdsourcing, we limited evaluation to a sample of 1000 conversations and used primary systems only. 
All systems were assigned the same conversations.
Each output was rated by 3 randomly-assigned judges provided by a crowdsourcing service. 
Judges were asked to rate outputs in context for Relevance and Interest using a 5-point Likert scale. 
Not unexpectedly, the constant baseline performed moderately well on Relevance (2.60), but poorly on Interest judgments, where it was statistically indistinguishable from the (low) random baseline (random: 2.35, constant: 2.32).
The best system returned a composite score of 2.93 (Relevance: 2.99, Interest: 2.87).
This remains well below the human baseline of 3.55 (Relevance: 3.61, Interest: 3.49). 
After replacing spammers, interrater agreement on a converted 3-way scale was fair, with Fleiss' Kappa at 0.39 for Relevance and 0.38 for Interest.

\subsection{Summary}
\label{sec:task2:Summary}
The sentence generation task challenged participants to produce interesting and informative end-to-end conversational responses that drew on textual background knowledge.
In this respect, the task was significantly more challenging that the DSTC6 task that was focused on the conversational dimensions of response generation.
In general, competing system outputs were judged by humans to be more relevant and interesting than our constant and random baselines.
It is also clear, however, that the quality gap between human and system responses is substantial, indicating that there is considerable space for research in future algorithmic improvements.

\section{Audio Visual Scene-aware dialog Track}
\label{avsd}

In this track, we consider a new research target: a dialog system that can discuss dynamic scenes with humans.
This lies at the intersection of research in natural language processing, computer vision, and audio processing.
As described above, end-to-end dialog modeling using paired input and output sentences has been proposed as a way to reduce the cost of data preparation and system development.
Such end-to-end approaches have been shown to better handle flexible conversations by enabling model training on large conversational datasets \cite{vinyals2015neural, hori2018overview}.
However, current dialog systems cannot understand a scene and have a conversation about what is going on in it.
To develop systems that can carry on a conversation about objects and events taking place around the machines or the users, the systems need to understand not only a dialog history but also the video and audio information in the scene.
In the field of computer vision, interaction with humans about visual information has been explored in {\em visual question answering} (VQA) by \cite{VQA} and {\em visual dialog} by \cite{visdial_rl}.
These tasks have been the focus of intense research, aiming to (1) generate answers to questions about things and events in a single static image and (2) hold a meaningful dialog with humans about an image using natural, conversational language in an end-to-end framework.
To capture the semantics of dynamic scenes, recent research has focused on {\em video description}.
The state-of-the-art in video description uses multimodal fusion
%attention mechanism that selectively attends to 
to combine different input modalities (feature types), such spatiotemporal motion features and audio features
%, in addition to temporal attention
 proposed by ~\cite{hori2017attention}. 
%To generate system responses based on a dialog context and an audiovisual scene, we need to combine context aware dialog technologies and audio visual semantic technologies.
Since the recent revolution of neural network models allows us to combine different modules into a single end-to-end differentiable network, this framework allow us to build scene aware dialog systems by combining end-to-end dialog and multimodal video description approaches.
We can simultaneously use video features and user utterances as input to an encoder-decoder-based system whose outputs are natural-language responses.

\subsection{Task definition}
In this track, the system must generate responses to a user input in the context of a given dialog.
The dialog context consists of a dialog history between the user and the system in addition to the video and audio information in the scene.
There are two tasks, each with two versions (a and b):
\begin{description} 
  \item[Task 1: Video and Text] (a) Using the video and text training data provided but no external data sources, other than publicly available pre-trained feature extraction models (b) Also using external data for training.

  \item[Task 2: Text Only] (a) Do not use the input videos for training or testing. Use only the text training data (dialogs and video descriptions) provided. (b) Any publicly available text data may be used for training.
\end{description}

\subsection{Data}
\label{avsd-data}
To set up the Audio Visual Scene-Aware Dialog (AVSD) track,
%we collected text-based human dialog data for videos used for human action recognition datasets.  
we collected text-based dialogs about short videos of Charades by ~\cite{sigurdsson2016hollywood}\footnote{\url{http://allenai.org/plato/charades/}}, a dataset of untrimmed and multi-action videos, along with video descriptions in \cite{alamri2018audio}. 
The data collection paradigm for dialogs was similar to the one described in~\cite{DBLP:journals/corr/DasKGSYMPB16}, in which for each image, two parties interacted via a text interface to yield a dialog.
In~\cite{DBLP:journals/corr/DasKGSYMPB16}, each dialog consisted of a sequence of questions and answers about an image.
In the video scene-aware dialog case, two parties had a discussion about events in a video.
One of the two parties played the role of an answerer who had already watched the video.
The answerer answered questions asked by their counterpart -- the questioner.
The questioner was not allowed to watch the whole video but were able to see the first, middle and last frames of the video as single static images.
The two had 10 rounds of QA, in which the questioner asked about the events that happened between the frames.
At the end, the questioner summarized the events in the video as a description.

The DSTC7 AVSD official dataset contains 7,659, 1,787 and 1,710 dialogs for training, validation and testing, respectively.
The questions and answers of the AVSD dataset mainly consists of 5 to 8 words, making them longer and more descriptive than VQA.
The dialog contains questions asking about objects, actions and audio information in the videos.
Although we tried to collect questions directly relevant to the event displayed, some questions ask about abstract information in the video such as how to begin the videos and the duration of the videos.
Table~\ref{tab:sample} shows an example dialog from the data set.

\begin{table}[h]
\centering
\caption{\footnotesize{An example dialog from the AVSD dataset.}}
\label{tab:sample}
\small
\begin{tabular}{|c|l|l|}
\hline
  & Questioner	&  Answerer \\
\hline
\hline
QA1 & What kind of room does this appear to be? & He appears to be in the bedroom. \\
\hline
QA2 & How does the video begin? & By him entering the room.\\
\hline
QA3 & Does he have anything in his hands?	& He pick up a towel and folds it.\\
\hline
QA4 & What does he do with it ? & He just folds them and leaves them on the chair.\\
\hline
QA5 & What does he do next? & Nothing much except this activity.\\
\hline
QA6 & Does he speak in the video?	& No he did not speak at all.\\
\hline
QA7 & Is there anyone else in room at all? & No he appears alone there.\\
\hline
QA8 & Can you see or hear any pets in the video?	& No pets to see in this clip.\\
\hline
QA9 & Is there any noise in the video of importance?	& Not any noise important there.\\
\hline
QA10 & Are there any other actions in the video?	&  Nothing else important to know.\\
\hline
\end{tabular}
\end{table}

\if 0
%%-------------------------
\begin{table}[h]
\centering
\caption{\footnotesize{The dialog data for the DSTC7 AVSD track. The test videos for this challenge were partially selected from the official test data of the Charades challenge.}}
\label{tab:data}
\begin{tabular}{ll|ccc}
\hline
& 
& training 
& validation 
& test \\
\hline
\# of dialogs 
& 
& 7,659
& 1,787
& 1,710 \\
\# of turns   
& 
& 153,180
& 35,740
& 13,490
\\
\# of words   
& 
&  1,450,754
&  339,006
&  110,252
\\
\hline
\end{tabular}
\end{table}
%%-------------------------
\fi 

\subsection{Evaluation}
In this challenge, the quality of a system's automatically generated sentences is evaluated using objective measures.
These determine how similar the generated responses are to ground truths from humans and how natural and informative the responses are.
To collect more possible answers in response to the questions for the test videos, we asked 5 humans to watch a video and read a dialog between a questioner and an answer about the video, and then to generate an answer in response to the question.
We evaluated the automatically generated answers by comparing with the 6 ground truth sentences (one original answer and 5 subsequently collected answers).
We used the MSCOCO evaluation tool for objective evaluation of system outputs \footnote{https://github.com/tylin/coco-caption}.
The supported metrics include word-overlap-based metrics such as BLEU, METEOR, ROUGE\_L, and CIDEr.

We also collected human ratings for each system response using a 5 point Likert Scale, where humans rated system responses given a dialog context as: 5 for Very good, 4 for Good, 3 for Acceptable, 2 for Poor, and 1 for Very poor.
Since we the dataset contains questions and answers,
%using from an existing video description dataset,  
%Charades~\cite{sigurdsson2016hollywood}, for Dialog System Technology Challenge the %7th edition (DSTC7)\footnote{http://workshop.colips.org/dstc7/call.html}. 
we asked humans to consider correctness of the answers and also naturalness, informativeness, and appropriateness of the response according to the given context.

\subsection{Results}
The AVSD Task received 31 system submission from 9 teams.
%Systems are listed as team\_M(N), where M is the team index and N is an identifier for a particular system submitted by that team. 
%``Ext. Data'' in the table denotes whether or not the system used external data for training and/or testing, where only team\_3(5) used external data (web data) for response generation.
We built a baseline end-to-end dialog system that can generate answers in response to user questions about events in a video sequence as described in \cite{hori2018end}.
Our architecture is similar to the Hierarchical Recurrent Encoder in~\cite{DBLP:journals/corr/DasKGSYMPB16}.
The question, visual features, and the dialog history are fed into corresponding LSTM-based encoders to build up a context embedding, and then the outputs of the encoders are fed into an LSTM-based decoder to generate an answer.
The history consists of encodings of QA pairs. 
We feed multimodal attention-based video features into the LSTM encoder instead of single static image features.
%Figure \ref{fig:scene-aware-dialog} shows the architecture of our video scene-aware dialog system.
The systems submitted deployed LSTM, BLSTM, and GRU with cross entropy as the objective function.
The best system applied "Hierarchical and Co-Attention mechanisms to combine text and vision" from \cite{libovicky2017attention,lu2016hierarchical}.
Table \ref{tab:result4avsd} shows the evaluation results for the baseline and best systems.
Under this evaluation, the human rating for the original answers was 3.938.
%-------------------------
\begin{table}[h]
\centering 
\caption{\footnotesize{Performance comparison between the baseline and the best system.}}
\label{tab:result4avsd}
\begin{tabular}{c|c|c|c|c}
\hline
System & BLEU-4 & METEOR & CIDEr & Human rating\\
\hline
\hline
Baseline &  0.309 &	0.215 & 0.746 & 2.848\\
\hline
Best & 0.394 & 0.267 & 1.094 & 3.491\\ 
\hline
\end{tabular}
\end{table}

%Hierarchical and Co-Attention mechanisms to combine text and vision + Fine-tuned a model trained on the How2 data used for JSALT 2018 Grounded Sequence-to-Sequence Transduction\footnote{https://srvk.github.io/jsalt-2018-grounded-s2s/} with the DSTC7 data.

\subsection{Summary}
We introduced a new challenge task and dataset for Audio Visual Scene-Aware Dialog (AVSD) in DSTC7.
This is the first attempt to combine end-to-end conversation and end-to-end multimodal video description models into a single end-to-end differentiable network to build scene-aware dialog systems.
The best system applied hierarchical attention mechanisms to combine text and visual information, improving by 22\% over the human rating for the baseline system.  
%The best system used 
%%"Hierarchical and Co-Attention mechanisms to combine text and vision"
%"Hierarchical attention for Multimodal Dialog Systems", improving by %22\% over the human rating for the baseline system.  
The language models trained from QA are still strong approaches and the power to predict the objects and events in the video is not sufficient to answer the questions correctly.
Future work includes more detailed analysis of the correlation between the QA text and the video scenes.

\section{Conclusion and Future Directions}
\label{conclusions}

In this paper, we summarized tasks conducted on the seventh dialog system technology challenge (DSTC7): sentence selection, sentence generation, and audio visual scene-aware dialog.
The sentence selection track contained several variations on the response selection problem, with five sub-tasks and two new datasets.
The sentence generation track provided a test of knowledge-grounded response production, with the aim of creating more controllable generators.
The audio visual scene-aware track raised a new problem in which dialog is generated about a given video in a variety of sub-tasks.

All of the data described in this paper will be provided as a large-scale benchmark of dialog systems from several viewpoints, after the challenge, to support future dialog system research.
However, there are several major remaining challenges for dialog systems. 
For example, transferring models trained on large-scale data-sets to a variety of domains that do not have enough data is a known issue for dialog systems, as mentioned in DSTC3. 
Data created this challenge, which focused on end-to-end learning, does not address this issue, which would require expanding to a larger variety of domains.
We expect to continue the challenge in the future, providing new testbeds that work towards the remaining open problems of dialog system research. 

%\bibliography{references,refs_generation,refs_AVSD}

\begin{thebibliography}{29}
\providecommand{\natexlab}[1]{#1}
\providecommand{\url}[1]{\texttt{#1}}
\expandafter\ifx\csname urlstyle\endcsname\relax
  \providecommand{\doi}[1]{doi: #1}\else
  \providecommand{\doi}{doi: \begingroup \urlstyle{rm}\Url}\fi

\bibitem[Alamri et~al.(2018)Alamri, Cartillier, Lopes, Das, Wang, Essa, Batra,
  Parikh, Cherian, Marks, et~al.]{alamri2018audio}
H.~Alamri, V.~Cartillier, R.~G. Lopes, A.~Das, J.~Wang, I.~Essa, D.~Batra,
  D.~Parikh, A.~Cherian, T.~K. Marks, et~al.
\newblock Audio visual scene-aware dialog (avsd) challenge at dstc7.
\newblock \emph{arXiv preprint arXiv:1806.00525}, 2018.

\bibitem[Antol et~al.(2015)Antol, Agrawal, Lu, Mitchell, Batra, Zitnick, and
  Parikh]{VQA}
S.~Antol, A.~Agrawal, J.~Lu, M.~Mitchell, D.~Batra, C.~L. Zitnick, and
  D.~Parikh.
\newblock {VQA}: {V}isual {Q}uestion {A}nswering.
\newblock In \emph{International Conference on Computer Vision (ICCV)}, 2015.

\bibitem[Chen et~al.(2016)Chen, Zhu, Ling, Wei, Jiang, and
  Inkpen]{chen2016enhanced}
Q.~Chen, X.~Zhu, Z.~Ling, S.~Wei, H.~Jiang, and D.~Inkpen.
\newblock Enhanced lstm for natural language inference.
\newblock \emph{arXiv preprint arXiv:1609.06038}, 2016.

\bibitem[Das et~al.(2016)Das, Kottur, Gupta, Singh, Yadav, Moura, Parikh, and
  Batra]{DBLP:journals/corr/DasKGSYMPB16}
A.~Das, S.~Kottur, K.~Gupta, A.~Singh, D.~Yadav, J.~M.~F. Moura, D.~Parikh, and
  D.~Batra.
\newblock Visual dialog.
\newblock \emph{CoRR}, abs/1611.08669, 2016.
\newblock URL \url{http://arxiv.org/abs/1611.08669}.

\bibitem[Das et~al.(2017)Das, Kottur, Moura, Lee, and Batra]{visdial_rl}
A.~Das, S.~Kottur, J.~M. Moura, S.~Lee, and D.~Batra.
\newblock Learning cooperative visual dialog agents with deep reinforcement
  learning.
\newblock In \emph{International Conference on Computer Vision (ICCV)}, 2017.

\bibitem[Doddington(2002)]{Doddington:2002}
G.~Doddington.
\newblock Automatic evaluation of machine translation quality using n-gram
  co-occurrence statistics.
\newblock In \emph{Proceedings of the Second International Conference on Human
  Language Technology Research}, HLT '02, pages 138--145, 2002.

\bibitem[Ghazvininejad et~al.(2018)Ghazvininejad, Brockett, Chang, Dolan, Gao,
  Yih, and Galley]{grounded2018}
M.~Ghazvininejad, C.~Brockett, M.~Chang, B.~Dolan, J.~Gao, W.~Yih, and
  M.~Galley.
\newblock A knowledge-grounded neural conversation model.
\newblock \emph{AAAI}, 2018.

\bibitem[Henderson et~al.(2014{\natexlab{a}})Henderson, Thomson, and
  Williams]{henderson2014second}
M.~Henderson, B.~Thomson, and J.~D. Williams.
\newblock The second dialog state tracking challenge.
\newblock In \emph{Proceedings of the 15th Annual Meeting of the Special
  Interest Group on Discourse and Dialogue (SIGDIAL)}, pages 263--272,
  2014{\natexlab{a}}.

\bibitem[Henderson et~al.(2014{\natexlab{b}})Henderson, Thomson, and
  Williams]{henderson2014third}
M.~Henderson, B.~Thomson, and J.~D. Williams.
\newblock The third dialog state tracking challenge.
\newblock In \emph{Spoken Language Technology Workshop (SLT), 2014 IEEE}, pages
  324--329. IEEE, 2014{\natexlab{b}}.

\bibitem[Hori and Hori(2017)]{DSTC6T2}
C.~Hori and T.~Hori.
\newblock End-to-end conversation modeling track in {DSTC6}.
\newblock \emph{arXiv:1706.07440}, 2017.

\bibitem[Hori et~al.(2017)Hori, Hori, Lee, Zhang, Harsham, Hershey, Marks, and
  Sumi]{hori2017attention}
C.~Hori, T.~Hori, T.-Y. Lee, Z.~Zhang, B.~Harsham, J.~R. Hershey, T.~K. Marks,
  and K.~Sumi.
\newblock Attention-based multimodal fusion for video description.
\newblock In \emph{ICCV}, 2017.

\bibitem[Hori et~al.(2018{\natexlab{a}})Hori, Alamri, Wang, Winchern, Hori,
  Cherian, Marks, Cartillier, Lopes, Das, et~al.]{hori2018end}
C.~Hori, H.~Alamri, J.~Wang, G.~Winchern, T.~Hori, A.~Cherian, T.~K. Marks,
  V.~Cartillier, R.~G. Lopes, A.~Das, et~al.
\newblock End-to-end audio visual scene-aware dialog using multimodal
  attention-based video features.
\newblock \emph{arXiv preprint arXiv:1806.08409}, 2018{\natexlab{a}}.

\bibitem[Hori et~al.(2018{\natexlab{b}})Hori, Perez, Higasinaka, Hori, Boureau,
  Inaba, Tsunomori, Takahashi, Yoshino, and Kim]{hori2018overview}
C.~Hori, J.~Perez, R.~Higasinaka, T.~Hori, Y.-L. Boureau, M.~Inaba,
  Y.~Tsunomori, T.~Takahashi, K.~Yoshino, and S.~Kim.
\newblock Overview of the sixth dialog system technology challenge: Dstc6.
\newblock \emph{Computer Speech \& Language}, 2018{\natexlab{b}}.

\bibitem[Kim et~al.(2016)Kim, D'Haro, Banchs, Williams, Henderson, and
  Yoshino]{kim2016fifth}
S.~Kim, L.~F. D'Haro, R.~E. Banchs, J.~D. Williams, M.~Henderson, and
  K.~Yoshino.
\newblock The fifth dialog state tracking challenge.
\newblock In \emph{Spoken Language Technology Workshop (SLT), 2016 IEEE}, pages
  511--517. IEEE, 2016.

\bibitem[Kim et~al.(2017)Kim, D’Haro, Banchs, Williams, and
  Henderson]{kim2017fourth}
S.~Kim, L.~F. D’Haro, R.~E. Banchs, J.~D. Williams, and M.~Henderson.
\newblock The fourth dialog state tracking challenge.
\newblock In \emph{Dialogues with Social Robots}, pages 435--449. Springer,
  2017.

\bibitem[Kummerfeld et~al.(2018)Kummerfeld, Gouravajhala, Peper, Athreya,
  Gunasekara, Ganhotra, Patel, Polymenakos, and
  Lasecki]{Kummerfeld-EtAl:2018:arXiv}
J.~K. Kummerfeld, S.~R. Gouravajhala, J.~Peper, V.~Athreya, C.~Gunasekara,
  J.~Ganhotra, S.~S. Patel, L.~Polymenakos, and W.~S. Lasecki.
\newblock Analyzing assumptions in conversation disentanglement research
  through the lens of a new dataset and model.
\newblock \emph{ArXiv e-prints}, October 2018.
\newblock URL \url{https://arxiv.org/pdf/1810.11118.pdf}.

\bibitem[Lavie and Agarwal(2007)]{Lavie2007}
A.~Lavie and A.~Agarwal.
\newblock {METEOR}: An automatic metric for mt evaluation with high levels of
  correlation with human judgments.
\newblock In \emph{Proc. of the Second Workshop on Statistical Machine
  Translation}, StatMT '07, pages 228--231, Stroudsburg, PA, USA, 2007.
  Association for Computational Linguistics.
\newblock URL \url{http://dl.acm.org/citation.cfm?id=1626355.1626389}.

\bibitem[Libovick{\`y} and Helcl(2017)]{libovicky2017attention}
J.~Libovick{\`y} and J.~Helcl.
\newblock Attention strategies for multi-source sequence-to-sequence learning.
\newblock \emph{arXiv preprint arXiv:1704.06567}, 2017.

\bibitem[Lowe et~al.(2015)Lowe, Pow, Serban, and Pineau]{lowe-EtAl:2015:W15-46}
R.~Lowe, N.~Pow, I.~Serban, and J.~Pineau.
\newblock The ubuntu dialogue corpus: A large dataset for research in
  unstructured multi-turn dialogue systems.
\newblock In \emph{Proceedings of the 16th Annual Meeting of the Special
  Interest Group on Discourse and Dialogue}, pages 285--294, Prague, Czech
  Republic, September 2015. Association for Computational Linguistics.
\newblock URL \url{http://aclweb.org/anthology/W15-4640}.

\bibitem[Lu et~al.(2016)Lu, Yang, Batra, and Parikh]{lu2016hierarchical}
J.~Lu, J.~Yang, D.~Batra, and D.~Parikh.
\newblock Hierarchical question-image co-attention for visual question
  answering.
\newblock In \emph{Advances In Neural Information Processing Systems}, pages
  289--297, 2016.

\bibitem[Papineni et~al.(2002)Papineni, Roukos, Ward, and
  Zhu]{Papineni2002BLEU}
K.~Papineni, S.~Roukos, T.~Ward, and W.-J. Zhu.
\newblock {\sc Bleu}: a method for automatic evaluation of machine translation.
\newblock \emph{ACL}, 2002.

\bibitem[Ritter et~al.(2011)Ritter, Cherry, and Dolan]{ritter2011data}
A.~Ritter, C.~Cherry, and W.~B. Dolan.
\newblock Data-driven response generation in social media.
\newblock \emph{EMNLP}, 2011.

\bibitem[Serban et~al.(2016)Serban, Sordoni, Bengio, Courville, and
  Pineau]{serban2015hierarchical}
I.~V. Serban, A.~Sordoni, Y.~Bengio, A.~Courville, and J.~Pineau.
\newblock Building end-to-end dialogue systems using generative hierarchical
  neural network models.
\newblock \emph{AAAI}, 2016.

\bibitem[Shang et~al.(2015)Shang, Lu, and Li]{shang2015neural}
L.~Shang, Z.~Lu, and H.~Li.
\newblock Neural responding machine for short-text conversation.
\newblock \emph{ACL-IJCNLP}, 2015.

\bibitem[Sigurdsson et~al.(2016)Sigurdsson, Varol, Wang, Laptev, Farhadi, and
  Gupta]{sigurdsson2016hollywood}
G.~A. Sigurdsson, G.~Varol, X.~Wang, I.~Laptev, A.~Farhadi, and A.~Gupta.
\newblock Hollywood in homes: Crowdsourcing data collection for activity
  understanding.
\newblock \emph{ArXiv}, 2016.
\newblock URL \url{http://arxiv.org/abs/1604.01753}.

\bibitem[Sordoni et~al.(2015)Sordoni, Galley, Auli, Brockett, Ji, Mitchell,
  Nie, Gao, and Dolan]{sordoni2015}
A.~Sordoni, M.~Galley, M.~Auli, C.~Brockett, Y.~Ji, M.~Mitchell, J.-Y. Nie,
  J.~Gao, and B.~Dolan.
\newblock A neural network approach to context-sensitive generation of
  conversational responses.
\newblock \emph{NAACL-HLT}, 2015.

\bibitem[Vedantam et~al.(2015)Vedantam, Zitnick, and Parikh]{vedantam:15}
R.~Vedantam, C.~L. Zitnick, and D.~Parikh.
\newblock {CIDEr}: Consensus-based image description evaluation.
\newblock In \emph{CVPR}, pages 4566--4575, 2015.

\bibitem[Vinyals and Le(2015)]{vinyals2015neural}
O.~Vinyals and Q.~Le.
\newblock A neural conversational model.
\newblock \emph{ICML}, 2015.

\bibitem[Williams et~al.(2013)Williams, Raux, Ramachandran, and
  Black]{williams2013dialog}
J.~Williams, A.~Raux, D.~Ramachandran, and A.~Black.
\newblock The dialog state tracking challenge.
\newblock In \emph{Proceedings of the SIGDIAL 2013 Conference}, pages 404--413,
  2013.

\end{thebibliography}

\end{document}